\documentclass[letterpaper, 10 pt, conference]{ieeeconf}  

\IEEEoverridecommandlockouts                              

\usepackage{amsmath}
\usepackage{booktabs}
\usepackage{graphicx}
\usepackage{subcaption}
\usepackage{comment}
\usepackage{dblfloatfix}
\usepackage{tabularx}
\usepackage{bm}
\usepackage{esvect}  
\usepackage{amssymb}
\usepackage{balance}

\usepackage{xcolor}

\begin{document}
\title{Beyond Success: Quantifying Demonstration Quality in Learning from Demonstration}

\author{Muhammad Bilal$^{a}$, Nir Lipovetzky$^{a}$, Denny Oetomo$^{b}$, and Wafa Johal$^{a}$
\thanks{$^{a}$Muhammad Bilal, Nir Lipovetzky, and Wafa Johal are with the School of Computing and Information Systems, The University of Melbourne, Parkville, Victoria 3010, Australia (email: \{m.bilal; nir.lipovetzky; wafa.johal\}@unimelb.edu.au)}
\thanks{$^{b}$Denny Oetomo is with the Department of Mechanical Engineering, The University of Melbourne,
        Parkville, Victoria 3010, Australia (email: doetomo@unimelb.edu.au)}
\thanks{This work was partially supported by the Australian Research Council (Grant No. DE210100858)}
}

\maketitle
\thispagestyle{empty}
\pagestyle{empty}

\begin{abstract}

Learning from Demonstration (LfD) empowers novice users to teach robots daily life tasks without writing sophisticated code, thereby promoting the democratization of robotics. However, novice users often provide sub-optimal demonstrations, which can potentially impact the robot's ability to efficiently learn and execute the tasks. Prior research has assessed the quality of demonstrations by evaluating the robot's task performance; however, the approach remains insufficient to qualify individual demonstrations, leaving the reason for classifying demonstrations as high- or low-quality unknown. Therefore, this simulation-based study aims to quantify the quality of individual demonstration at each step by incorporating motion-related quality features such as manipulability and joint-space jerk. To assess the efficacy of these features, we initially evaluated the given demonstrations---taking into account each quality feature---to rank them from high- to low-quality. Subsequently, we investigated the impact of demonstration's quality on task performance and the quality of task execution. In this pursuit, we trained a series of LfD models for distinct manipulation tasks: cube lifting and pick-and-place of soda can. Our results illustrate a strong correlation between ranked demonstrations and the quality of task execution. Interestingly, we observed that the quality features have a significant impact on task performance, particularly when the provided demonstrations exhibit diversity in terms of quality. Overall, this analysis enables quantifying the quality of individual demonstrations based on motion-related quality features, thus improving learning from demonstration.
\end{abstract}
\begin{figure*}[!t]
    \centering
    \includegraphics[width=0.98\textwidth, trim=0cm 1.5cm 0cm 0.6cm, clip]{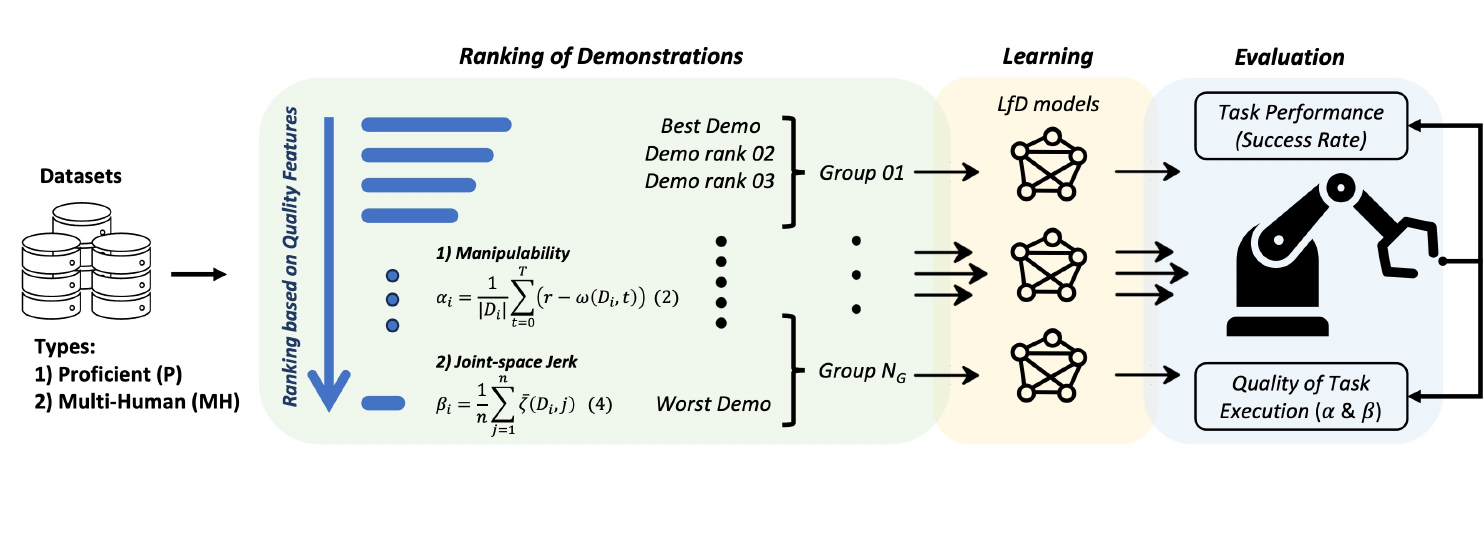}
    \caption{\textbf{Overview} The given datasets have been evaluated based on the quality features, including manipulability and joint-space jerk, and ranked them from high to low quality. At next step, we trained an LfD model for each group, where $N_G$ denotes the total number of groups, followed by rollouts the learned LfD model in order to compute the task performance and assess the quality of task execution based on the quality features.}
    \vspace{-15pt} 
    \label{fig:overview}
\end{figure*}
\vspace{-10pt}
\vspace{0.08in}
\section{Introduction}
To overcome the obstacle of using robots for daily life tasks without programming skills, the `Learning from Demonstration' (LfD) approach offers a promising solution. This strategy, inspired by human interactions, enables novice users to instruct robots through demonstrations instead of navigating the complexities of coding \cite{Argall2009}. Through LfD, users can seamlessly impart tasks to robots, making the integration of robotic assistance more accessible and user-friendly for individuals without programming expertise. \par
One of the key goals in LfD is to enable a robotic system to efficiently learn and execute tasks with a minimal set of demonstrations. Achieving this goal is heavily contingent upon the quality of the provided demonstrations as well as modality of demonstration \cite{Ravich2020}, as it directly shapes the overall competency of the robotic system. According to Laskey et al. \cite{Laskey2017}, LfD algorithms fall into two broad categories: robot-centric and human-centric. In the former type, the robotic system initiates its learning process from a given set of demonstrations and then actively learns from its surroundings, incorporating feedback provided by the demonstrator \cite{Ross2011}. On the other side, the human-centric approach involves learning an LfD model from a set of demonstrations provided by novice users; an example of this approach is imitation learning \cite{chernova2014}. In the literature, researchers have devoted considerable attention to the robot-centric side, with comparatively less emphasis on the human-centric aspect for enhancing LfD \cite{osa2018}. 
In the domain of LfD, the \emph{quality of demonstration} encompasses the dexterity and smoothness exhibited by novice users \cite{Ravich2020}. Meanwhile, the \emph{quality of task execution} gauges the robot's aptitude to carry out the intended tasks, taking into consideration features such as manipulability \cite{vahrenkamp2015} and joint-space jerk \cite{chen2003}. This dual focus on both the quality of demonstration and the quality of task execution significantly contributes to the enhanced performance and adaptability of the robotic system within the LfD paradigm. 

\indent Existing evaluation of demonstration's quality mainly focuses on appraising the robot's task performance, classifying demonstrations as either high-quality or low-quality based on the success or failure of the task \cite{sakr2021, sakr2022}. While quality assessment of task performance is crucial for human-centric LfD, previous approaches do not allow for assessment of individual demonstration. Instead, the task performance's evaluation is typically applied to a set of demonstrations utilized for training an LfD model---a machine learning model that reproduces actions observed during demonstrations---employing task performance as the primary metric. The inclusion of a learning phase to train the LfD model makes this process potentially time-intensive. Furthermore, the underlying reason for the given demonstrations being of high or low quality remains unclear to the user. In response to these limitations, it becomes crucial to systematically measure and assess each demonstration, taking into consideration various quality features. Feature-based methods \cite{dhal2022, duboue2020} present a promising approach to address the mentioned limitations, specifically the issues of being time-intensive and lacking clarity on why a demonstration is classified as high or low quality. Therefore, this study aims to investigate a set of quality features by quantifying individual demonstrations at each step based on quality features and subsequently assessing the impact of the demonstration's quality on task performance and the quality of task execution.
\par
This study makes three primary contributions: firstly, it advocates for the utilisation of motion-based features to quantify the quality of individual demonstrations at an early stage and regardless of learning an LfD model. Secondly, it highlights the transferability of quality features from provided demonstrations to the corresponding LfD model generated motions. Finally, it posits that our quality features can serve as an indicator of task performance when used in a diverse set of demonstrations.  
\vspace{-3pt}
\section{Background and Related Work}
\subsection{Assessing Novice Users' Input in LfD}
Several LfD algorithms assume that users are proficient and can provide optimal demonstrations \cite{Ravich2020}. However, this assumption often neglects the heterogeneous nature of human demonstrators, leading to sub-optimal demonstrations \cite{Amershi2014}. The quality of demonstrations can vary from user to user,  influenced significantly by their understanding of the task and their capabilities to provide effective demonstrations \cite{Calinon2007}. \par
Pais et al. \cite{Pais2015} proposed three metrics to evaluate the user’s ability, including maneuvering the tool, consistency in teaching, and arm coordination. 
The metric such as consistency in teaching may not inherently depict the quality of a demonstration, as users could potentially exhibit similar errors across different demonstrations. Fischer et al. \cite{Fischer2016} compared teaching modalities in LfD, and used three common issues associated with novice operators: 1) self-collision, 2) singularity, and 3) excessive force on the end-effector. However, multiple ways of controlling the robot have been studied but did not evaluate the impact of these issues on the learned LfD model. \par
To evaluate the diversity of demonstrations across task space, Sena et al. \cite{sena2020} proposed two metrics: 1) teaching efficacy and 2) teaching efficiency. 
These metrics alone are insufficient for assessing the quality of individual demonstrations, as they primarily address the issue of data sparsity. Furthermore, the concept of entropy has been utilized to identify areas where additional demonstrations are required for effective robot learning \cite{sakr2023can}. The entropy approach is an interesting but focuses on diversity of the demonstrations instead of quantification of individual demonstrations at each step. 
\begin{figure}[!t]
  \centering
  \begin{subfigure}[b]{0.47\columnwidth}
    \includegraphics[width=\linewidth]{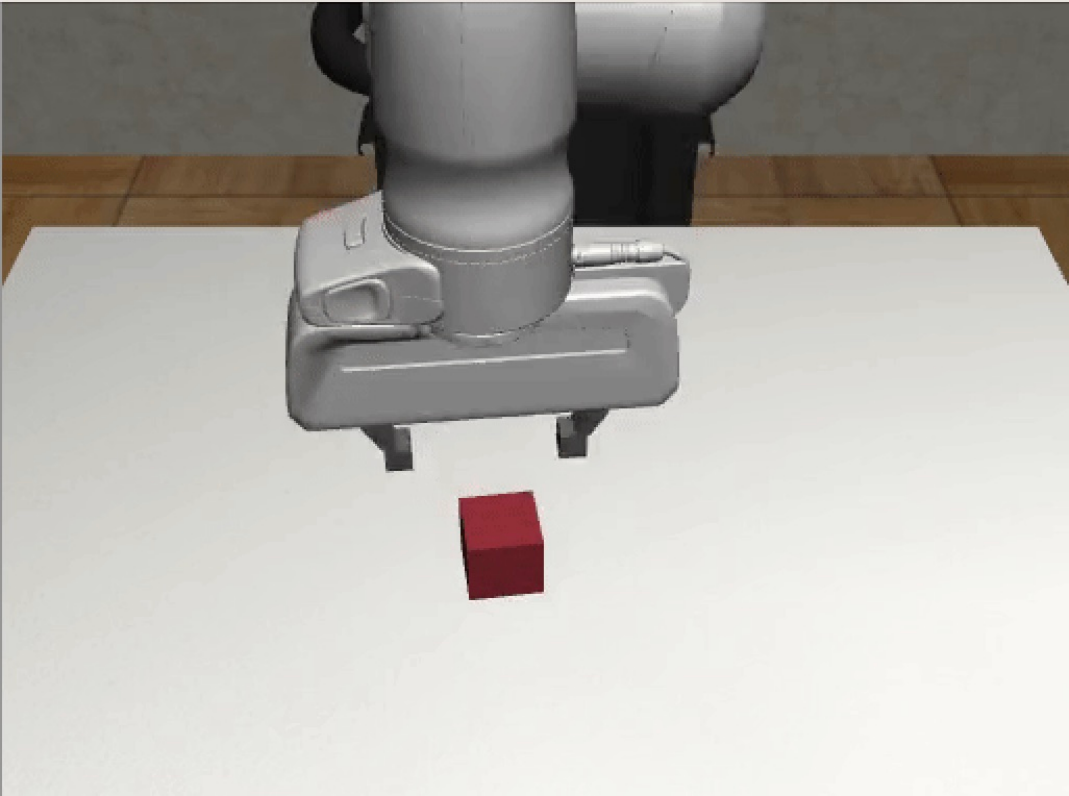}
    \subcaption{Lift task}
  \end{subfigure}
  \hfill
  \begin{subfigure}[b]{0.47\columnwidth}
    \includegraphics[width=\linewidth]{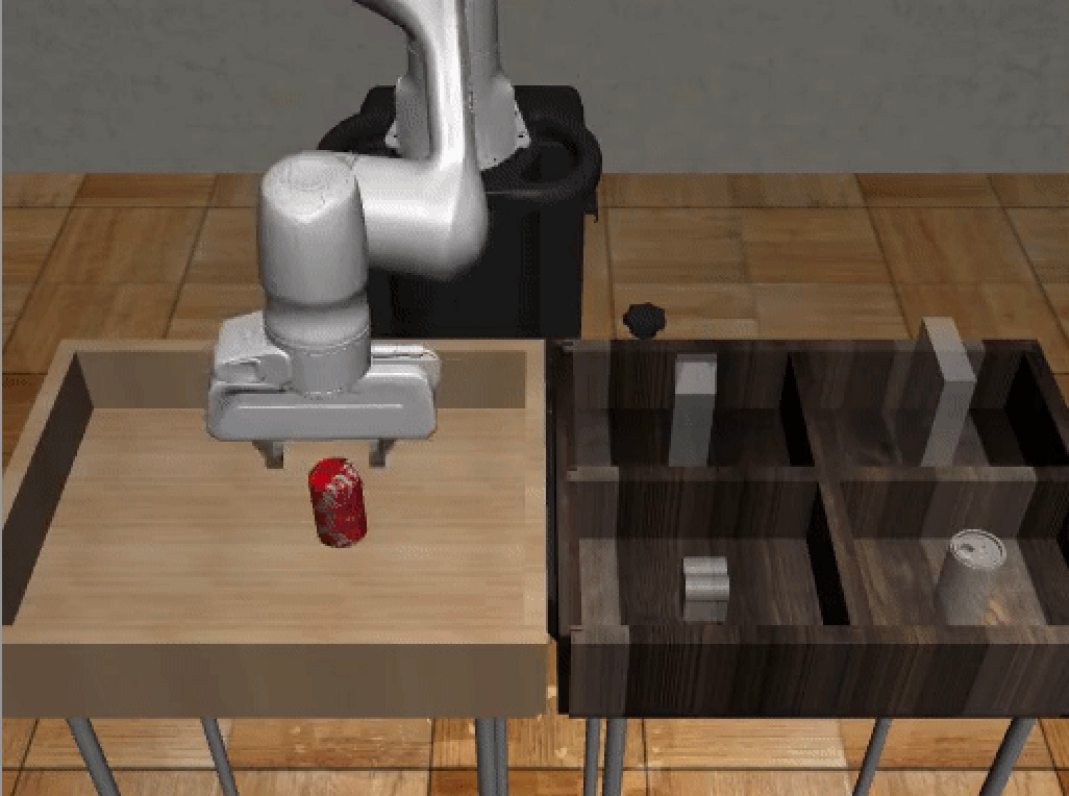}
    \subcaption{Pick-and-place of soda can}
  \end{subfigure}
  \caption{Manipulation tasks using Robosuite platform \cite{robosuite2020}}
   \vspace{-10pt} 
  \label{fig:images}

\end{figure}
\subsection{Assessing Quality of Task Execution}
To quantify the quality of task execution, previous research collected six core quality features: \emph{manipulability}, \emph{jerk} at Cartesian and joint-space, \emph{trajectory length} in Cartesian and joint-space, and \emph{robot’s joint limits} \cite{sakr2021}. Nevertheless, these features have not been utilized for quantifying the quality of demonstrations. Among these quality features, Chen et al. \cite{chen2003} investigated the importance of smoothness in robot learning, where they proposed an approach to identify and eliminate noises by providing alternative optimal control commands to the robot. Secondly, the manipulability is an important feature to consider because it directly impacts the maneuverability of the robotic system \cite{vahrenkamp2015}. Furthermore, Jaquier et al. \cite{Jaquier2021} proposed an interesting approach to learn and replicate manipulability ellipsoids from provided demonstrations. However, we aim to investigate the impact of quality features irrespective of learning particular feature from given demonstrations. \par
Our work introduces innovation by explicitly assessing individual demonstration based on quality features and evaluating their influence on both task performance and the quality of task execution. We selected two quality features—manipulability and joint-space jerk—based on the relationships established by
 \cite{Ravich2020}. To clarify, high-quality demonstrations are those with the highest measures of these features, while low-quality demonstrations have the lowest measures. We hypothesize: \\
 H1 The quality of demonstration correlates with task performance. \\
 H2 The quality of demonstration correlates with the quality of task execution.
\section{Features Formulation}
In this section, we present the mathematical formulation of motion-related quality features, including manipulability and joint-space jerk. Additionally, we provide a detailed description of the ranking criteria applied to the given demonstrations.

\subsection{Manipulability} 
\textit{Manipulability measure} quantifies how far the current configuration of a robotic system is from a singular configuration, where the robotic system cannot move in a specific direction \cite{yoshikawa1985}. We selected the manipulability measure because it directly represents the robot's dexterity. In assessing the demonstrations, higher measures of manipulability correspond to high-quality demonstrations, while the opposite holds for lower measures of manipulability. \par
Consider a robot with $n$ degrees of freedom whose joint variables are denoted by $\vec{q}=\langle q_0, q_1, \ldots, q_n\rangle$, where $q_j \in \mathbb{R}$. A demonstration $(D)$ can be defined as a joint-space trajectory $D=\langle \vec{q}_0, \ldots, \vec{q}_T\rangle$ over the time period $0 \leq t \leq T$, where $\vec{q}_t$ and $T$ represent the joint configuration of the robot at time step $t$ and the total duration of the demonstration, respectively. \par
Let $\mathcal{D}=\langle D_1, D_2, \ldots, D_K \rangle$ depict the set of all given demonstrations, where $D_i$ indicates the $i^{th}$ demonstration. The total number of demonstrations is represented by $K$, where $K=|\mathcal{D}|$. For a task with $m$ dimensions, the Jacobian matrix $J(\vec{q}): \mathbb{R}^{n} \rightarrow \mathbb{R}^{m \times n}$ is computed for the given joint configuration $(\vec{q})$ of the robot. The manipulability measure $(\omega)$ for demonstration $D_i$ at time step $t$ is then expressed as: 
\begin{align}
    \omega(D_i,t) &=\sqrt{det \ J\left(\vec{q}_{t}\right)J^{'}\left(\vec{q}_{t}\right)}
\end{align}
Finally, we computed the manipulability based metric $\alpha_i \in \mathbb{R}$ for demonstration $D_i$, by finding the average difference between the manipulability measure at each time step and the global maximum value $r = \{max{\left\langle \vec{\omega}(D_1), \vec{\omega}(D_2), \ldots,\vec{\omega}(D_K) \right \rangle}\} \in \mathbb{R}$, where $\vec{\omega}(D_i)=\langle \omega(D_i,0),\ldots,\omega(D_i,T)\rangle$.
\begin{align}\label{eq:manip}
    \alpha_{i} &= \frac{1}{|D_i|} \displaystyle\sum_{\substack{t=0}}^{T} \left(r - \omega(D_i,t)\right)
\end{align}
A higher $\alpha$ value indicates a poorer manipulability measure, representing a low quality demonstration, and vice versa. 
\subsection{Joint-space Jerk}
For a robot joint index $j$ in a given demonstration $D_i$, the jerk $\zeta (D_i,j,t)$ is computed by taking the triple derivative of the robot's joint position for the given period $0 \leq t \leq T$.

Next, the mean jerk value $\overline{\zeta}(D_i,j) \in \mathbb{R}$ is calculated followed by computing a jerk based metric $\beta_i \in \mathbb{R}$ for demonstration $D_i$, indicating how much demonstration $D_i$ contains jerky motion at joint-space level.

\begin{align}
        \overline{\zeta}(D_i,j)= \frac{1}{|D_i|} \sum_{\substack{t=0}}^{T} \zeta (D_i,j,t)
\end{align}
\begin{align}\label{eq:jerk}
        \beta_{i}= \frac{1}{n} \sum_{\substack{j=1}}^{n} {\overline{\zeta}}(D_i,j)
\end{align}
Based on the $\beta$ value, ranking is determined with lower values indicating smoother demonstrations, while higher values correspond to a higher magnitude of jerks, indicating low quality demonstrations.
\section{Analyses Setup}
In this section, we discuss the dataset and the procedure employed for computing task performance. Additionally, we elaborate on the criteria for selecting an algorithm for training LfD models, focusing on task performance while utilizing a minimum number of demonstrations. 
\subsection{Dataset}
In this study, we used the robomimic v0.1 dataset for the analyses \cite{Mandlekar2021, robomimic}. The demonstrations in the dataset were collected using the RoboTurk tool \cite{mandlekar2018}. Among multiple tasks, we considered the lift and pick and place tasks because they encapsulate fundamental manipulation skills such as grasping, lifting, and placement as shown in Fig. \ref{fig:images}. To investigate the influence of diverse-quality demonstrations, we considered both a ``proficient'' (P) and a ``multi-human'' (MH) dataset, with a total of 200 and 300 successful demonstrations for each manipulation task, respectively. In the multi-human dataset, the demonstrations were gathered by six operators with varying expertise, comprising two proficient, two normal, and two worst performers. Each individual contributed a total of 50 demonstrations, resulting in a dataset with mixed quality.  On the other side, the proficient dataset consists of demonstrations provided by expert user. 
\subsection{Performance Criteria}
To compute the success rates, we utilized the learned LfD models---obtained using one of the LfD algorithms---to execute the same task as observed during the demonstrations. In both the demonstration and execution phases, the initial pose of the object was set randomly. For the lift task, the learned model requires to successfully lift a cube to a specific height of $0.06 m$, while for the pick-and-place of soda can, the robot needs to pick up a soda can and accurately place it in its designated location (see Figure \ref{fig:images}b). 

\subsection{Learning Algorithms}
In real-world scenarios, it is often impractical for human users to provide a large number of demonstrations; therefore, one of the objectives in LfD is to train a model with a minimum number of demonstrations \cite{chen2020active}. In this pursuit, we propose to identify which of the algorithm provided by Mandlekar et al. \cite{Mandlekar2021}  exhibits the best performance with a minimal number of demonstrations. We test: Behavioral Cloning (BC) \cite{pomerleau1988}, Hierarchical Behavioural Cloning (HBC) \cite{mandlekar2020learning}, Batch-Constrained Q-Learning (BCQ) \cite{fujimoto2019}, and Implicit Reinforcement without Interaction at Scale (IRIS) \cite{mandlekar2020iris}. \par
The BC algorithm executes simple regression over a sequence of state-action pairs in the given dataset. Several variants of BC algorithm, including BC based on Gaussian mixture model (GMM), have been investigated for different types of manipulation tasks \cite{Mandlekar2021}. The HBC and IRIS algorithms are identical, except the latter includes the addition of a value function in a high-level mechanism to locate a state with the maximum expectation. In this way, the IRIS algorithm is suitable for identifying near optimal state-action pair. Lastly, the BCQ algorithm is an offline batch reinforcement learning \cite{fujimoto2019}. In this study, we considered an offline learning, where the algorithms are not permitted to consider additional samples.  
\par
For each algorithm, we computed the success rates --- representing the number of successful task executions divided by the total number of rollouts \cite{Mandlekar2021} --- for each set of demonstrations. The term “rollout” represents the deployment of the learned LfD model on a robotic system to evaluate its performance on the given task. To determine the minimal number of demonstrations to achieve an acceptable success rate, we devised 10 groups of varying sizes ranging from 10 to 100 demonstrations, with an incremental factor of 10 demonstrations. Additionally, we devised one group which consist of 5 demonstrations. Overall, this resulted in training 44 models per task, with 11 models for each algorithm.
To validate the effectiveness of the models' performance, we employed the k-fold cross-validation technique for each group. 
\begin{figure}[!t]
  \centering
  \begin{subfigure}[b]{0.8\columnwidth}
    \includegraphics[width=\linewidth]{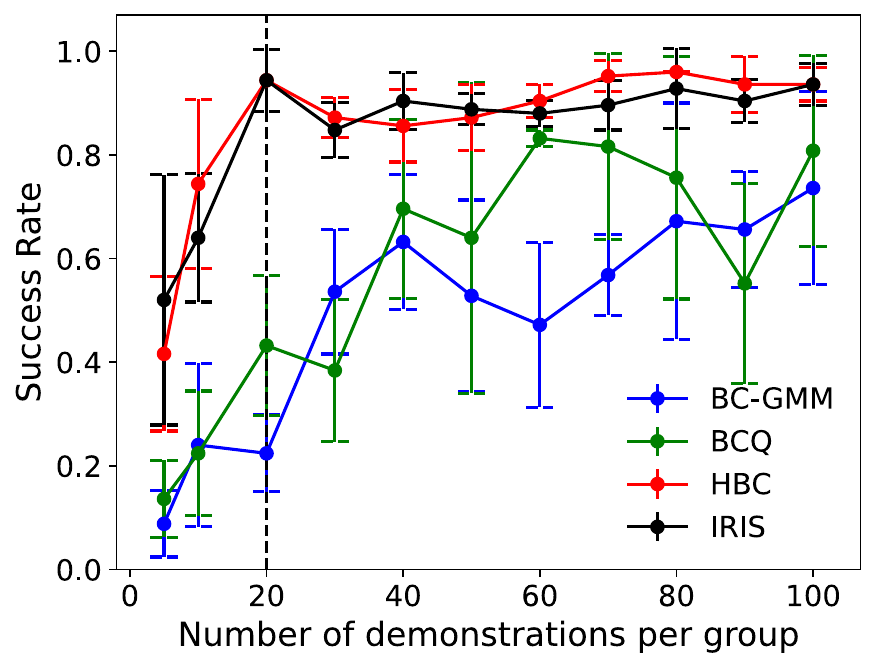}
    \caption{Lift task}
    \label{fig:algo_perf_lift}
  \end{subfigure}
  \hfill
  \begin{subfigure}[b]{0.8\columnwidth}
    \includegraphics[width=\linewidth]{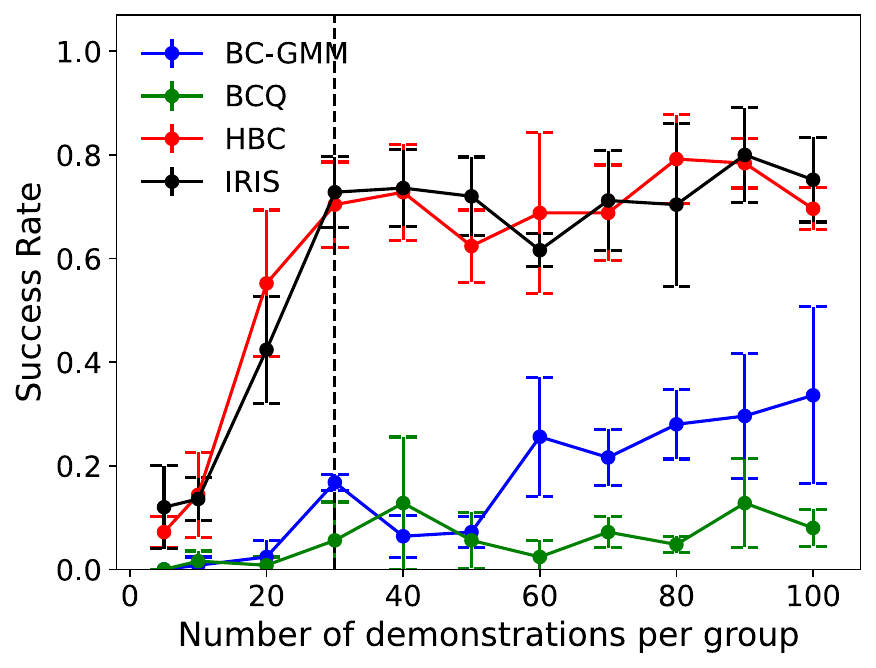}
    \caption{Pick-and-place of soda can}
    \label{fig:algo_perf_can}
  \end{subfigure}
  \caption{This figure illustrates the performance of algorithms for the given different number of demonstrations per group, considering both the (a) lift task and (b) pick-and-place of soda can.}
  \label{fig:tasks}
\end{figure}
\subsection{Experiment Parameters}
Observing the learning curves \cite{Mandlekar2021}, a suitable number of epochs for both the lift and pick-and-place tasks is $250$, utilizing a batch size of $100$. Next, the learned model was subjected to $25$ rollouts after every $50$ epochs, allowing the computation of success rates. To ensure the adaptability of the learned model to an actual robotic system, the rollouts were performed on the same robotic system using the Robosuite platform \cite{robosuite2020}. Throughout the experiments, the demonstration data was partitioned into training and validation sets with a split ratio of $80\%$ and $20\%$, respectively.

\begin{figure*}[t]
    \centering
    \includegraphics[width=\textwidth]{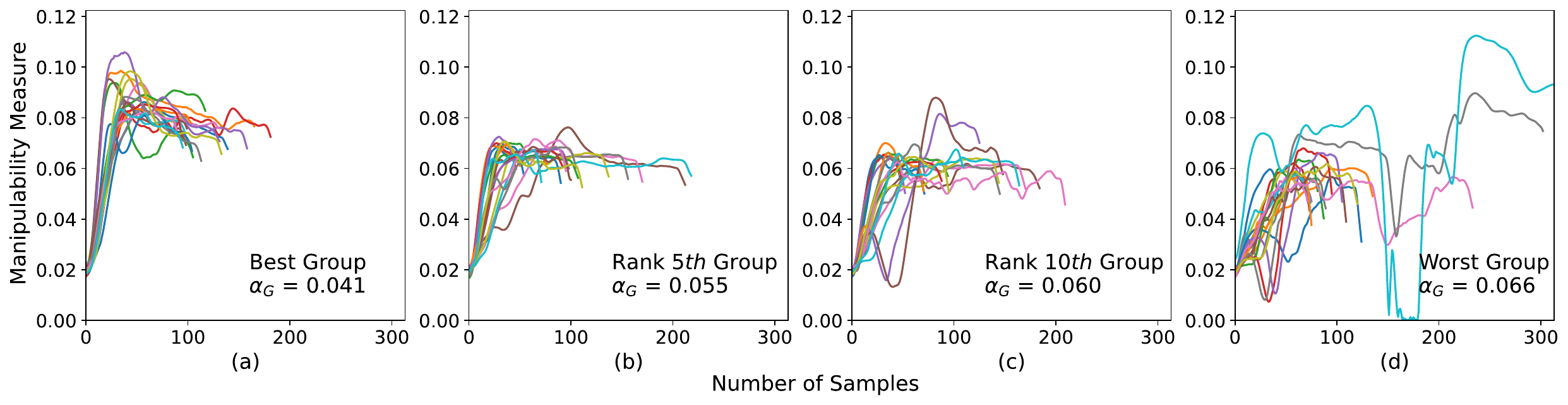}
    \caption{Each sub-figure illustrates ranked demonstrations based on the manipulability feature, where $\alpha_G$ represents the average of manipulability based metrics $(\alpha)$ of the corresponding group. In Fig. \ref{fig:input-data-manip}d, the worst demonstration (cyan line) in the dataset falls to zero values for some interval, indicating a singular configuration.}
    \label{fig:input-data-manip}
\end{figure*}
\begin{figure*}[b]
    \centering
    \includegraphics[width=\textwidth]{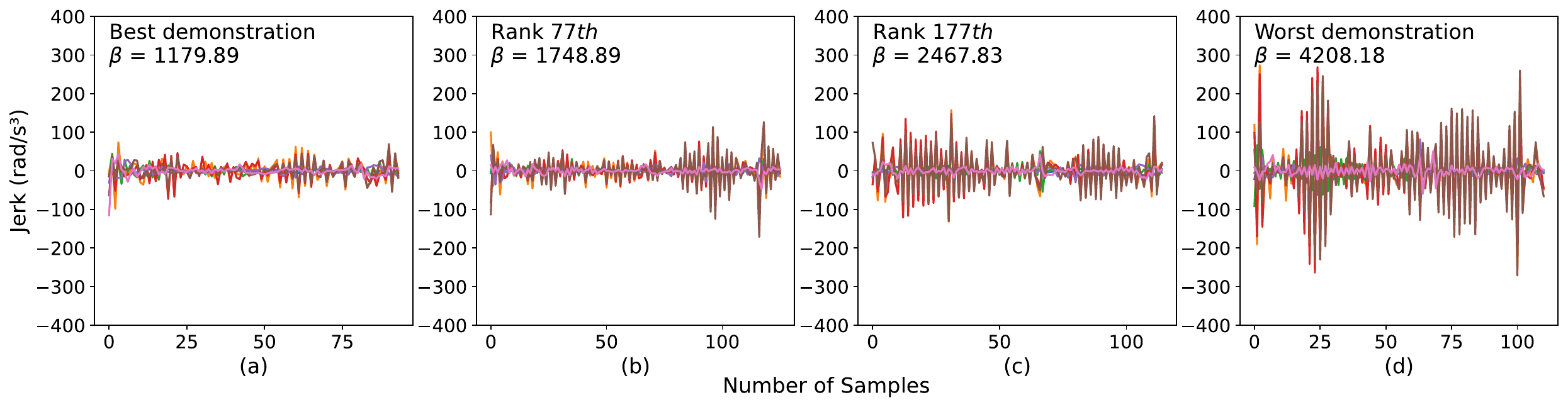}
    \caption{Each sub-figure depicts a ranked demonstration based on the jerk feature. The worst demonstration exhibits jerky motion throughout the entire length of the demonstration, while the best demonstration indicates smoother motion profiles.}
    \label{fig:input-data-jerk}
\end{figure*}

\section{Selection of Learning Algorithm} \label{sec:selection}
In our preliminary analyses, the algorithm with the minimum number of demonstrations required to achieve an acceptable success rate of $90\%$ and $70\%$, values observed from the learning curves where the algorithm converged \cite{mandlekar2020learning}, for the lift and pick-and-place tasks, respectively. For each algorithm, the models were trained on the same set of demonstrations using 5-fold cross-validation with same performance criteria and experimental parameters. Additionally, the multi-human dataset was used for both manipulation tasks: lift task and pick-and-place of soda can. \par
\vspace{2pt}
Regarding BC-GMM algorithm, the success rates are significantly low for a small number of demonstrations; however, it improves with the increasing number of demonstrations as shown in Fig. \ref{fig:algo_perf_lift}. Each algorithm, as illustrated in Fig. \ref{fig:algo_perf_can}, exhibited lower success rates as the task complexity increased from the simple lift task to the relatively complex pick-and-place of soda can. For example, the success rates of BC-GMM algorithm are almost halved for the pick-and-place task as compared to the lift task. In addition, the BCQ algorithm also exhibited poor task performance, less than $0.2$ for all the given sets of demonstrations as depicted in Fig. \ref{fig:algo_perf_can}. In the case of HBC and IRIS algorithms, the success rates showed satisfactory outcomes for both manipulation tasks as shown in Fig. \ref{fig:algo_perf_lift} and \ref{fig:algo_perf_can}. \par
\vspace{2pt}
A notable drop in task performance was observed, particularly in the case of the BCQ algorithm for the pick-and-place of soda can. One possible explanation is that the BCQ algorithm extrapolates to unseen states to find a solution, making it challenging to accomplish complex manipulation tasks. Similar issue has been addressed for grid world problems as well \cite{mandlekar2020iris}. From the results, we noticed that the IRIS algorithm requires more data as the task complexity increases to yield satisfactory results. One possible reason is that the IRIS algorithm, due to its limitation \cite{mandlekar2020iris}, demands every possible state of the object in the training dataset. Therefore, for complex manipulation tasks, more data is needed to cover as many states as possible in the case of IRIS algorithm. \par
\vspace{2pt}
Overall, according to the criterion of achieving the maximum success rate, the minimum number of demonstrations is $20$ per group for the lift task, as shown in Fig. \ref{fig:algo_perf_lift}, while the minimum number is $30$ per group for the relatively complex pick-and-place of soda can, as illustrated in Fig. \ref{fig:algo_perf_can}. Both algorithms, HBC and IRIS, exhibited satisfactory performance. Given the application of the IRIS algorithm in handling diverse and suboptimal demonstrations \cite{mandlekar2020iris}, we selected the IRIS algorithm for our analyses.  
\begin{table*}
\begin{center}
  \caption{Statistics of dataset based on quality features along with correlation coefficients and the corresponding p-values.}
  \label{tab:corr-coeff}
  \begin{tabular}{lcccccc}
    \toprule
     && &\multicolumn{2}{c}{Lift task}&\multicolumn{2}{c}{Pick-and-Place Soda Can}\\
  &&  &Proficient (P)& Multi-Human (MH)& Proficient (P)&Multi-Human (MH)\\
  \midrule
 Stats& Manip&  mean $\pm$ std & $0.032\pm0.004$& $0.057\pm0.007$& $0.035\pm0.009$&$0.042\pm0.011$\\
 & Jerk &  mean $\pm$ std & $685\pm212$& $\bm{1401\pm1256}$& $1940\pm513$&$\bm{3031\pm1868}$\\
    \midrule
     H1&Manip&      corr&$0.17$& $-0.11$& $-0.28$& $0.11$\\
 & &  p-value&$0.635$& $0.680$& $0.590$&$0.750$\\
     &Jerk&      corr&$0.38$& $\bm{-0.60}$& $0.31$& $\bm{-0.66}$\\
 & &  p-value&$0.270$& $\bm{0.019}$& $0.55$&$\bm{0.036}$\\
    \midrule
     H2&Manip&  corr&$0.68$& $0.78$& $0.77$&$0.60$\\
 & & p-value& $9.59\times10^{-24}$& $9.62\times10^{-30}$& $2.35\times10^{-15}$&$1.63\times10^{-12}$\\
     &Jerk&  corr&$0.75$& $0.84$& $0.86$&$0.73$\\
 & & p-value& $2.06\times10^{-32}$& $4.38\times10^{-66}$& $1.68\times10^{-29}$&$3.39\times10^{-16}$\\
\end{tabular}
\end{center}
\end{table*}
\section{Quality Features and Ranking of Demonstrations}
In this section, we present the method followed for a series of experiments aiming at testing our hypotheses --- high quality demonstrations will correspond to higher task performance (H1) and will attain an efficient task execution (H2). As a first step, we discuss here the ranking of demonstrations based on the quality features. 
For each demonstration, considering both manipulation tasks provided by proficient and multi-human, we computed the metrics based on manipulability and jerk features using (\ref{eq:manip}) and (\ref{eq:jerk}), respectively. Finally, we ranked all demonstrations from high to lower value, creating groups of $20$ and $30$ consecutively for both the lift and the pick-and-place of soda can tasks, respectively. The best group comprises top-ranked demonstrations while the worst group contains low-ranked demonstrations. To illustrate the quality diversity among the provided demonstrations, the mean and standard deviation based on the quality features are summarized in Tab. \ref{tab:corr-coeff}.

\subsection{Manipulability}
Considering the multi-human dataset, the manipulability curve $\vec{\omega}(D)$ of each demonstration for the lift task can be represented as shown in Fig. \ref{fig:input-data-manip}. The best group, as illustrated in Fig. \ref{fig:input-data-manip}a, comprises demonstrations with high manipulability measures, with average group $\alpha_G$ value $0.041$. The term $\alpha_G$ represents the average of manipulability based metrics $(\alpha)$ of the corresponding group. As the value of $\alpha_G$ increases, the corresponding demonstrations in the group gradually fall to the lower manipulability measures as shown in Fig. \ref{fig:input-data-manip}. The worst group consists of demonstrations with average group $\alpha_G$ value $0.066$, including demonstration that contain zero values for some interval, representing a singular configuration, as shown in Fig. \ref{fig:input-data-manip}d.

\subsection{Joint-space Jerk}
For the pick-and-place of soda can using proficient dataset, the jerky motion of multiple demonstrations with varying magnitude are depicted in Fig. \ref{fig:input-data-jerk}. The best demonstration showcases the lowest jerk magnitude for all joints of the robotic system, as illustrated in Fig. \ref{fig:input-data-jerk}a, with a $\beta$ value 1179.89. Furthermore, the value of $\beta$ increases with the jerk magnitude, where the worst demonstrations exhibiting high magnitude with $\beta$ equals to 4208.18, as illustrated in Fig. \ref{fig:input-data-jerk}d.
\begin{figure*}[t]
    \centering
    \includegraphics[width=1\textwidth]{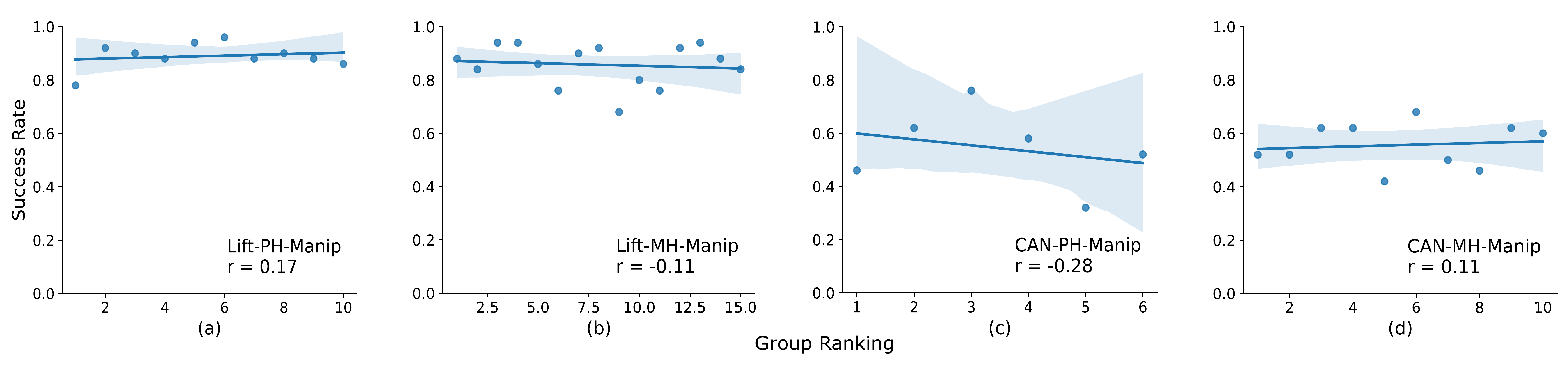}
    \caption{Correlation between input ranked groups based on the manipulability feature and the corresponding success rates}
    \label{fig:tp-corr-manip}
\end{figure*}
\begin{figure*}[t]
    \centering
    \includegraphics[width=1\textwidth]{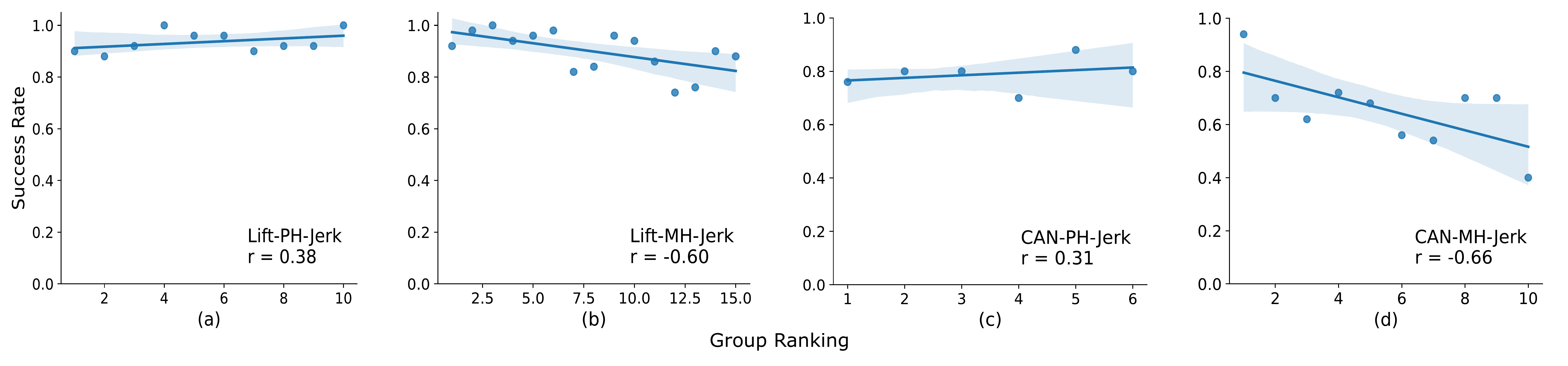}
    \caption{Correlation between input ranked groups based on the jerk feature and the corresponding success rates}
    \label{fig:tp-corr-jerk}
\end{figure*}
\begin{figure*}[t]
    \centering
    \includegraphics[width=1\textwidth]{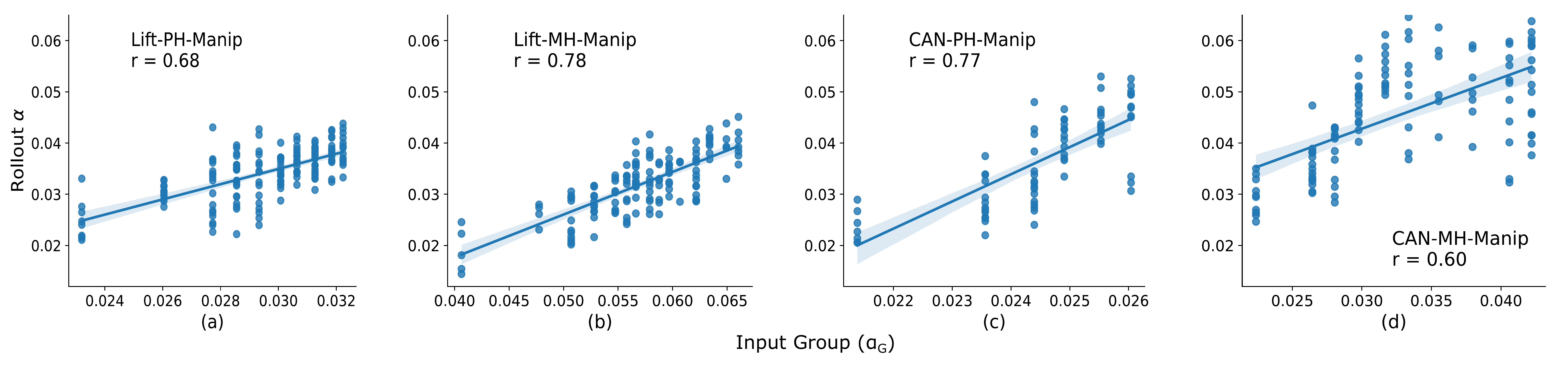}
    \caption{Correlation between input ranked groups based on the manipulability feature and the corresponding rollouts.}
    \label{fig:corr-manip}
\end{figure*}

\begin{figure*}[h!]
    \centering
    \includegraphics[width=1\textwidth]{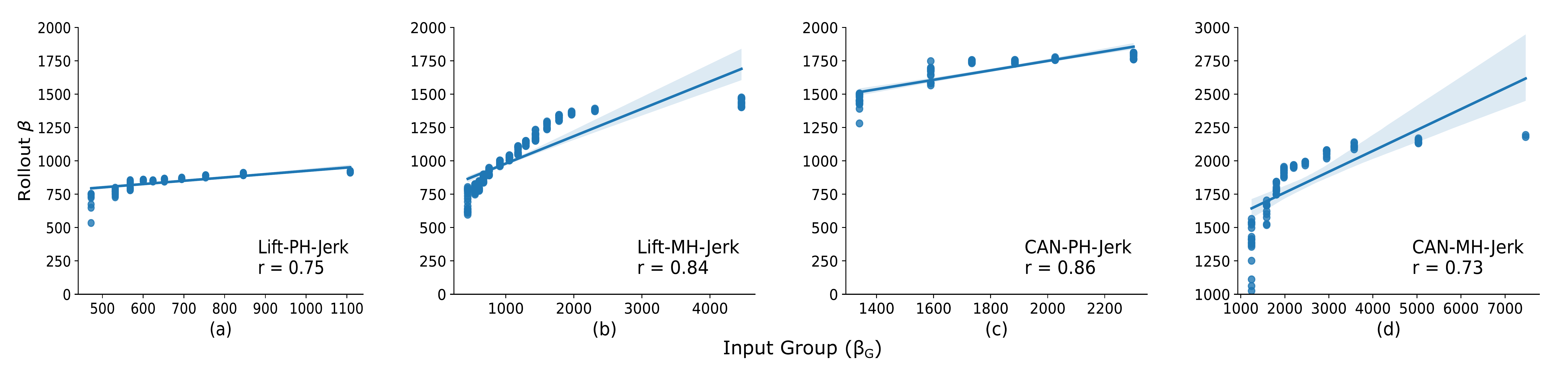}
    \caption{Correlation between input ranked groups based on the jerk feature and the corresponding rollouts.}
    \label{fig:corr-jerk}
\end{figure*}
\section{Results}
As per the results on the performance of various LfD algorithms on the lift and pick-and-place tasks (see Section \ref{sec:selection}), we choose to use IRIS in the following experiments.  
We trained a series of IRIS-LfD models for both manipulation tasks and analyzed the correlation between input ranked groups and corresponding task performance (H1), as well as the quality of task execution (H2). The findings of each study are presented in detail in the following subsections.
\subsection{Study 01 (H1)}
To evaluate the first hypothesis, we computed the correlation coefficients between ranked groups and the corresponding success rates (i.e. task performance). As depicted in Fig. \ref{fig:tp-corr-manip}, there is no apparent impact of the manipulability feature on task performance in both manipulation tasks as well as both datasets (P and MH). Moreover, the p-values exceed 0.05 in all cases. Regarding the jerk feature, the correlation coefficients also show no effect on task performance in the case of the proficient (P) dataset as shown in Fig. \ref{fig:tp-corr-jerk}a and \ref{fig:tp-corr-jerk}c. However, we found strong correlations for both manipulation tasks, lift ($r=-0.60,\ p= 0.019$) and pick-and-place of soda can ($r=-0.66,\ p= 0.036$), in the dataset provided by multi-human (MH), as illustrated in Fig. \ref{fig:tp-corr-jerk}b and \ref{fig:tp-corr-jerk}d. Overall, the impact of demonstration's quality on task performance is significant when the demonstrations are of diverse-quality.

\subsection{Study 02 (H2)}
Our second hypothesis aimed at assessing if quality features were implicitly learned by our LfD algorithm.
To calculate the outcomes of task execution's quality, we assessed each rollout using the same metrics as above, $\alpha$ and $\beta$. 
We then tried to see if there was any correlation between the quality ranking of each group of demonstrations used as input and the task execution quality produced after rollout. Furthermore, we used the actual values to provide a more realistic interpretation of the outcomes. The correlation analyses were performed for each feature across both types of datasets: proficient and multi-human. \par
Regarding the manipulability feature, there were strong correlations between input ranked groups and the corresponding rollouts as presented in Tab. \ref{tab:corr-coeff}. For the jerk feature, the coefficients also showed strong correlations between input-ranked groups and the corresponding rollouts as shown in Tab. \ref{tab:corr-coeff}. As compared to the manipulability feature, the correlation coefficients concerning the jerk feature are higher in each scenario. The overall correlation coefficients and the corresponding p-values are summarised in Tab. \ref{tab:corr-coeff}.

\section{Discussion}

\subsection{Role of Diversity (H1)}
Using the two robotmimic v0.1 datasets (P and MH), we observed no distinct correlation between input ranked groups and the success rates; however, two cases showed strong correlation with respect to the jerk feature as shown in Fig. \ref{fig:tp-corr-jerk}b and \ref{fig:tp-corr-jerk}d. \\
\textit{Joint-Space Jerk:} Regarding the above two cases, one of the possible reasons is that the multi-human dataset includes a diverse-quality of demonstrations. For instance, in the proficient-human dataset, the coefficient of variation (CV) for the lift and pick-and-place tasks is $\textbf{30.94\%}$ and $\textbf{26.44\%}$, respectively. Conversely, in the multi-human dataset, the CV values for these manipulation tasks are substantially higher at \textbf{$\textbf{89.65\%}$} and $\textbf{61.62\%}$. Furthermore, the corresponding p-values are below $0.05$, specifically $\textbf{0.019}$ and $\textbf{0.036}$ as shown in Tab. \ref{tab:corr-coeff}, respectively.  \\
\textit{Manipulability:} The CV values for the lift and pick-and-place tasks using the proficient-human dataset are $\textbf{12.5\%}$ and $\textbf{25.71\%}$, while in the multi-human dataset, the corresponding CV values are $\textbf{12.8\%}$ and $\textbf{26.19\%}$, respectively. Because the datasets were collected using a simulator, the data exhibits a skewed distribution, particularly concerning the manipulability feature. \\
Overall, we hypothesize that H1 may hold if we have a diverse quality of demonstrations. This result somewhat aligns with \cite{sakr2022}, although their study considered a set of demonstrations rather than individual ones. For simple tasks, such as cube lifting, we expect minimal variations in manipulability on real robots, while noticeable joint-space jerk may occur due to the potential lack of smoothness from novice users. Due to dataset limitations, we plan to further investigate this hypothesis (H1) in our future work.\par

\subsection{Learning Quality Features (H2)}
Based on findings in study 02, the results support our second hypothesis (H2) as shown in Tab. \ref{tab:corr-coeff}. For each quality feature, the coefficients show strong correlations between input-ranked groups and the corresponding rollouts. All of the correlation coefficients are higher than $\textbf{0.60}$ in both manipulation tasks and types of datasets. While the p-value for each case is below 0.05 as shown in Tab. \ref{tab:corr-coeff}. \\
In the multi-human dataset, as depicted in Fig. \ref{fig:corr-jerk}b and \ref{fig:corr-jerk}d, outliers with input group $(\beta_G)$ values around $4500$ and $7500$ slightly impact the results for both manipulation tasks. Additionally, the figures illustrate consistency among ranked groups, except for the last ones. \\
From the results, we posit that quality features can be transferred from given demonstrations to the corresponding LfD model. Alternatively, a set of demonstrations with high manipulability measure and low jerk values will yield an LfD model that executes tasks with nearly the same metrics as the given demonstrations. 

\section{Conclusion}
In this work, we investigated the role of quality of individual demonstration in LfD. We quantified each demonstration's quality based on the quality features and then evaluated the impact of demonstration's quality on both task performance and task execution's quality. This exploration involved the training of a set of LfD models using the robomimic v0.1 dataset. In the preliminary analysis, we selected the IRIS algorithm based on its performance, requiring 20 demonstrations for the lift task and 30 for the pick-and-place of soda can. In study 01, we computed the correlation coefficients between input-ranked groups and their corresponding success rates. The findings revealed a significant correlation in two instances, specifically when the demonstrations exhibited diverse qualities. Conversely, no evident correlations emerged in the cases where the demonstrations displayed less diverse quality. Regarding study 02, the results exhibited a strong correlation between the quality of input-ranked group of demonstrations and the corresponding rollouts for each quality feature. \par   
This signify that not only the system can learn to perform the task (i.e. typically assessed through success rate) but also how to perform the task (through quality features).  
We hope that this work can lead to develop new ways to assess LfD models optimizing quality features rather than only looking at success rates. Furthermore, the proposed methodology for assessing the quality of demonstrations can be applied to a diverse range of tasks.

 Although this study focused on two potential quality features using the IRIS algorithm, we aim to assess additional relevant features and algorithms in future research. We plan to develop a feedback system to make capable novice users to provide demonstrations of desired quality. This study enables to identify the cut-off regions beyond which the quality of task execution becomes unsatisfactory as shown in Fig. \ref{fig:corr-manip} and \ref{fig:corr-jerk}. Eventually, a better execution function can be developed to guide novice users to provide effective demonstrations. 

\balance 
\bibliographystyle{IEEEtran}
\bibliography{references}

\end{document}